\documentclass{article}

\usepackage{microtype}
\usepackage{graphicx}
\usepackage{subfig}
\usepackage{booktabs} 
\usepackage{float}
\usepackage{url}
\usepackage{amsmath,amsfonts,bm}

\usepackage{hyperref}

\usepackage[accepted]{icml2021}

\icmltitlerunning{Specialized federated learning using a mixture of experts}

\begin{document}

\twocolumn[
\icmltitle{Specialized federated learning using a mixture of experts}



\icmlsetsymbol{equal}{*}

\begin{icmlauthorlist}
\icmlauthor{Edvin Listo Zec}{rise}
\icmlauthor{Olof Mogren}{rise}
\icmlauthor{John Martinsson}{rise}
\icmlauthor{Leon René Sütfeld}{rise}
\icmlauthor{Daniel Gillblad}{aise}
\end{icmlauthorlist}

\icmlaffiliation{rise}{RISE Research Institutes of Sweden}
\icmlaffiliation{aise}{AI Sweden}

\icmlcorrespondingauthor{Edvin Listo Zec}{edvin.listo.zec@ri.se}

\icmlkeywords{Machine Learning, ICML}

\vskip 0.3in
]



\printAffiliationsAndNotice{}  

\begin{abstract}
In federated learning, clients share a global model that has been trained on decentralized local client data. Although federated learning shows significant promise as a key approach when data cannot be shared or centralized, current methods show limited privacy properties and have shortcomings when applied to common real-world scenarios, especially when client data is heterogeneous. In this paper, we propose an alternative method to learn a personalized model for each client in a federated setting, with greater generalization abilities than previous methods. To achieve this personalization we propose a federated learning framework using a mixture of experts to combine the specialist nature of a locally trained model with the generalist knowledge of a global model. We evaluate our method on a variety of datasets with different levels of data heterogeneity, and our results show that the mixture of experts model is better suited as a personalized model for devices in these settings, outperforming both fine-tuned global models and local specialists.


\end{abstract}

\section{Introduction}

In many real-world scenarios, data is distributed over a large number of devices or across many organizations, due to privacy concerns or communication limitations. Federated learning is a framework that can leverage this data in a distributed learning setup. This allows for the use of all participating clients' compute power, with the added benefit of a large decentralized training data set, while enhancing privacy and data security.

\begin{figure}[H]
    \centering
    \includegraphics[width=\linewidth]{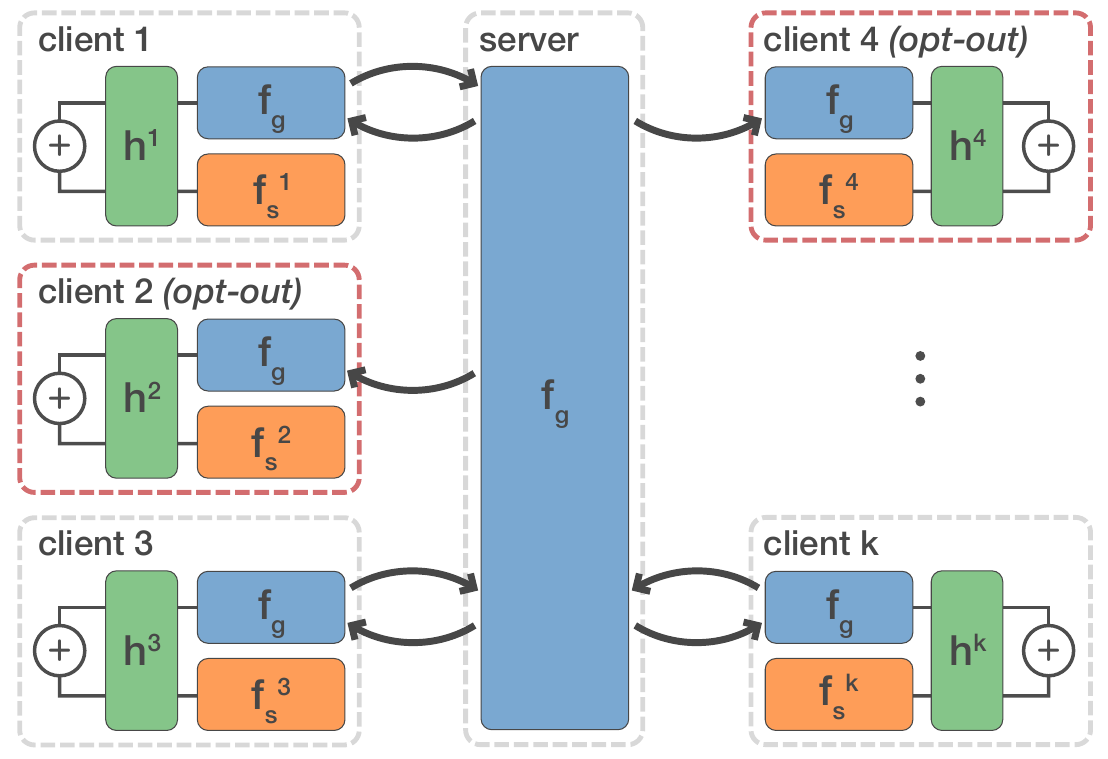}
    \caption{Federated mixtures of experts, consisting of a global model $f_g$ and local specialist models $f^k_s$ using local gating functions $h^k$. Some clients opt-out from federation, not contributing to the global model and keeping their data completely private.}
    \label{fig:overview}
\end{figure}

For instance, in keyboard prediction for smartphones, thousands or even millions of users produce keyboard input that can be leveraged as training data. The training can ensue directly on the devices, doing away with the need for costly data transfer, storage, and immense compute on a central server \citep{hard2018federated}. The medical field is another example area where data is often extremely sensitive and cannot be shared externally, thus requiring distributed and privacy-protecting approaches. 

The optimization problem that we solve in a federated learning setting is
\begin{equation}
     \min_{w\in\mathbb{R}^d} \frac{1}{K} \sum_{k=1}^K \mathbb{E}_{(x,y)\sim p_k} \left[ \ell_k (w ; \; x,y) \right]
\end{equation}
where $\ell_k$ is the loss for client $k$ and $(x,y)$ samples from the $k$th client's data distribution $p_k$. A central server is coordinating training between the $K$ local clients. The most prevalent algorithm for solving this optimization is the federated averaging (\textsc{FedAvg}) algorithm \citep{mcmahan2017communication}. In this solution, each client has its own client model, parameterized by $w^k$ which is trained on a local dataset for $E$ local epochs. When all clients have completed the training, their weights are sent to the central server where they are aggregated into a global model, parameterized by $w_g$. In \textsc{FedAvg}, the $k$ client models are combined by parameter averaging, weighted by the size of their respective local datasets:
\begin{equation}
w_g^{t+1} \leftarrow \sum_k \frac{n_k}{n} w^k_{t+1},
\end{equation}
where $n_k$ is the size of the dataset of client $k$ and $n=\sum_k n_k$. Finally, the new global model is sent out to each client, where it constitutes the starting point for the next round of (local) training. This process is repeated for a defined number of global communication rounds. 



The averaging of local models in parameter space generally works but requires some care to be taken in order to ensure convergence. \cite{mcmahan2017communication} showed that all local models need to be initialized with the same random seed for \textsc{FedAvg} to work. Extended phases of local training between communication rounds can similarly break training, indicating that the individual client models will over time diverge towards different local minima in the loss landscape. Similarly, different distributions between client datasets will also lead to divergence of client models \citep{mcmahan2017communication}. 

Depending on the use case, however, the existence of local datasets and the option to train models locally can be advantageous: specialized local models, optimized for the data distribution at hand may yield higher performance in the local context than a single global model, although typically at the cost of generalization performance. Keyboard prediction, for example, based on a global model may represent a good approximation of the population average, but could provide a better experience at the hands of a user when biased towards their individual writing style and word choices. This raises an important question -- when is a global FL-trained model better than a specialized local model? A specialist would be expected to perform better than a global generalist in a highly non-iid setting, whereas the global generalist would be expected to perform better in an iid setting.

There are several ways client distributions can be non-identical. The conditional distributions $P_i(x|y)$ on all clients $i$ may be the same, but the marginal distributions $P_i(x)$ may vary (covariate shift) or $P_i(y)$ may vary (prior probability shift). Further, if the marginal distribution $P(y)$ is the same on all clients, the conditional $P_i(x|y)$ may vary (same label, different features) or $P(x)$ is the same, but the conditional $P_i(y|x)$ varies (same features, different label). In this work we study non-identical distributions in the form of prior probability shift, although we hypothesize that our proposed method can handle other distributional shifts as well and would be an interesting direction for future work.

To address the issue of specialized local models within the federated learning setting, we propose a general framework based on a mixture of experts \citep{jacobs1991adaptive}. In this work we have one mixture of experts per client, each combining one local specialist model and one global model. Each client has a local \textit{gating function} that performs a weighting of the experts dependent on the input data. First, the global model is trained using \textsc{FedAvg}. This is followed by training of all clients' local specialist models, initialized with the trained federated global model. This is followed by training of the entire mixture, i.e., the local and global models as well as the gating function. A common problem with fine-tuned specialist models is that although they achieve better accuracy on local test data, they do not generalize as well as a global model. However, in our work we show that we can reach the same local accuracy on client data as a fine-tuned model, while retaining superior generalization performance.

While standard federated learning already shows some privacy enhancing properties, it has been shown that in some settings, properties of the client and of the training data may be reconstructed from the gradients communicated to the server \citep{wang2019beyond}. Therefore, we will work with a stronger notion of privacy in this paper. While existing solutions may be private enough for some settings, we will assume that clients requiring privacy for some of their data need this data not to have any influence on the training of the global model at all. Instead, our framework allows for a complete opt-out from the federation with some or all of the data for any client.
Clients with such preferences will still benefit from the global model and retain a high level of performance on their own, skewed data distribution. This is important when local datasets are particularly sensitive, as may be the case in medical applications.
Our experimental evaluations demonstrate the robustness of our learning framework with different levels of label heterogeneity in the data, and under varying fractions of opt-out clients.

\section{Related work}

Distributed machine learning
has been studied as a strategy to allow for training data to remain on the clients, giving it some aspects of privacy, while leveraging the power
of learning from bigger data and compute \citep{konevcny2016federated,shokri2015privacy,mcmahan2017communication,vanhaesebrouck2016decentralized,bellet2018personalized}.
The federated averaging algorithm \citep{mcmahan2017communication}
has been influential and demonstrated that averaging of the weights in neural network models
trained separately at the clients is successful in many settings, producing a federated
model that demonstrates the ability to generalize from limited subsets of data at the
clients. However, it has been shown that federated averaging struggles when data is not
independent and identically distributed among the clients, e.g., in the problem of prior probability shift. This illustrates the need for client personalization within federated learning \citep{kairouz2019advances,hsieh2020non}. 

In general, addressing class imbalance is still a relatively
understudied problem in deep learning \citep{johnson2019survey}.
A common approach for personalization on skewed label distributions is to first train a generalist model and then fine-tune it using more specific data. This approach is used in meta-learning \citep{finn2017model},
domain adaptation \citep{mansour2009domain}, and transfer learning \citep{oquab2014learning}.
For the distributed setting, fine-tuning was first proposed by
\cite{wang2019federated} who used federated averaging
to obtain a generalist model which was later fine-tuned
locally on each client, using each client's specific training data.
Some work has been inspired by the meta-learning paradigm to learn models that are specialized at the clients \citep{jiang2019improving,fallah2020personalized}.
\cite{arivazhagan2019federated} combined this strategy and ideas from transfer learning with deep neural networks and presented a solution where shallow layers are frozen, and the deeper layers are retrained at every client.

\cite{zhao2018federated} propose a strategy to improve training on non-iid client data by creating a subset of data which is globally shared between all clients. \cite{hsu2019measuring} show that performance degrades when client distributions shift, and propose to solve the problem via server momentum.
Recent strategies have also explored knowledge distillation techniques for federated learning \citep{jeong2018communication,he2020group,lin2020ensemble}, which show promising results in non-iid settings. 
%


\textbf{Mixing models.} \cite{deng2020adaptive} proposed to combine a global model $w$ trained using federated averaging, with a local model $v$ with a weight $\alpha_i$. To find optimal $\alpha_i$ they optimize $\alpha_{i}^{*}=\arg \min _{\alpha_{i} \in[0,1]} f_{i}\left(\alpha_{i} \boldsymbol{v}+\left(1-\alpha_{i}\right) \boldsymbol{w}\right)$ in every communication round.
While this weighting scheme will balance the two models, it is unable to adapt to
the strengths of the different members of the mix.

\cite{hanzely2020federated} proposed a solution that provides an explicit trade-off between global and local models by the introduction of an alternative learning scheme that does not take the full federation step at every round, but instead takes a step in the direction towards the federated average.

Mixture of experts have previously been used for learning private user models in a federated setting \citep{peterson2019private}. Although experiments are limited, the authors show that a mixture of a local and global model is more robust to differential privacy noise than a global model trained with federated averaging.


\textbf{Contributions.} In this work, we integrate mixture of experts into a non-iid federated setting in order to learn a personalized federated model for each client. More specifically, we leverage the strengths of a global model trained with federated averaging and a local model trained locally on each client. We show empirically on multiple datasets that our method outperforms both a locally trained model and a global federated model fine-tuned on test data from each client. Our results also show that our proposed method generalizes better than both baseline personalization methods. Further, we show our proposed method to be robust against low client participation, making it possible for clients to opt out from the global federation, and to keep the data completely private for these clients.

\section{Federated learning using a mixture of experts}
In this work, we present a framework for model personalization in a non-iid federated learning setting that builds on federated averaging and mixtures of experts. Our framework includes a personalized model for each client, which consists of a mixture of a globally trained model and a locally trained specialist. The local models never leave the clients, which gives strong privacy properties, while the global model is trained using federated averaging and leverages larger 
compute and data. In our framework, as illustrated in Figure \ref{fig:overview}, clients can choose to opt-out from the federation. This ensures complete privacy for their data, as no information from their data ever leaves the client.

Let $f_g$ be the global model with parameters $w_g$. We denote the index of clients by $k$ and the local specialist models by $f^k_s$ with parameters $w^k_s$. The gating function is denoted by $h^k$, parameterized with $w^k_h$. Training in the proposed framework is divided into three main parts. First, a global model $f_g$ is trained using federated averaging using opt-in data (see Section~\ref{sec:privacy-guarantees}). Second, a local specialist model $f_s^k$ is created for each client, initialized with the weights of the global model and fine-tuned using the local opt-in data on the client. Third, we train the mixture of experts 
\begin{equation}
    h^k(x)f^k_s(x) + (1-h^k(x))f_g(x).
\end{equation}
We freeze the weights of $f_g$, only updating the local specialist $f_s^k$ and the gating model $h^k$ on each client. By freezing the weights of $f_g$ in the last step, we ensure that the generalist knowledge in the model is not unlearned in the fine-tuning procedure.

As the mixture is trained, the two expert models $f_s$ and $f_g$ compete against each other, while the gating function sends an error signal guiding the winner of the two experts for every input. Over the course of this procedure, the gating function will learn to separate the input space given how well each expert perform on the task.

\subsection{Opting out from federation}
\label{sec:privacy-guarantees}
Users requiring high levels of privacy may not want to participate in the federation and disclose their locally trained model to a central server. For this reason, our proposed framework allows for clients to opt-out from federation.
Each client may arbitrarily partition its data into a part that is not used in the federation, and a part that is. No information from the opt-out data will ever leave the client. The system will still leverage learning from 
this data by using it to train the local specialist model $f_s^k$ and the gating model $h^k$.
This is a very flexible and useful property as it allows for the use of sensitive data in training of the private local models, while transformations of it, created by some privatization mechanism (e.g. differential privacy), can be used to train the federated model. 

Formally, each client dataset $\mathcal{D}^k$ is split into two non-overlapping datasets, $\mathcal{D}_{\mathcal{O}}^k$ and $\mathcal{D}_{\mathcal{I}}^k$, one of which has to be non-empty. The local model $f^{k}_{l}$ and the gating model $h^k$ is trained using the whole dataset $\mathcal{D}^k = \mathcal{D}_{\mathcal{O}}^k \cup \mathcal{D}_{\mathcal{I}}^k $, while the global model $f_g$ is trained with \textsc{FedAvg} using only the non-sensitive \textit{opt-in} dataset $\mathcal{D}_{\mathcal{I}}^k$. In Figure \ref{fig:overview} this is visualized by each client either opting-in or out all of its data. In our experiments, we assume that a client that opts out does so with its whole dataset, meaning that it puts all of its data in $\mathcal{D}_{\mathcal{O}}^k$.

\subsection{Optimization problem}
\textbf{Step 1: Train a global model.} We train the global model using \textsc{FedAvg}. In other words, globally we optimize 
\begin{equation}
\min_{w_g \in \mathbb{R}^d} \frac{1}{\left|\mathcal{D}_{\mathcal{I}}^k\right|} \sum_{k\in \mathcal{D}_{\mathcal{I}}^k} \mathbb{E}_{(x,y)\sim \mathcal{D}_{\mathcal{I}}^k} \left[ \ell_k (w_g ; \; x,y,\hat{y}_g) \right]
\end{equation}
for the opt-in dataset $\mathcal{D}_{\mathcal{I}}^k$. Here $\ell_k$ is the loss for the global model $w_g$ on client $k$ for the prediction $f_g(x) = \hat{y}_g$, and $\mathcal{D}_{\mathcal{I}}^k$ is the $k$th clients \textit{opt-in} data distribution.

\textbf{Step 2: Train local specialists.} The output model from \textsc{FedAvg} is fine-tuned on each clients datasets, minimizing the local loss. We initialize the specialist model with the global parameters $w_g$ and optimize:
\begin{equation}
  \min_{w_s^k \in \mathbb{R}^d} \mathbb{E}_{(x, y) \sim \mathcal{D}^k}\left[\ell_{k}(w^k_s ; \;  x, y,\hat{y}_l)\right] \quad \forall k=1,\dots,n.
\end{equation}
Here, $\ell_k$ is the loss for the prediction $\hat{y}_l = f^k_s(w^k_s; \; x)$  from the fine-tuned model on the input $x$ and $\mathcal{D}^k$ is the $k$th clients dataset.

\textbf{Step 3: Train local mixtures.} The local mixture of experts are trained using the gating models $h^k$, with the prediction error given by weighing the trained models $f_g$ and $f^k_s$:
\begin{equation}
\label{mixture}
\hat{y}_h =  h^k(x) f^k_s(x) + \left(1-h^k(x)\right) f_g(x) \quad \forall k = 1,\dots,n.
\end{equation}
In other words, at the end of a communication round, given $f^k_s$ and $f_g$, we optimize the mixture \eqref{mixture}:
\begin{equation}
 \min_{w_g, w^k_s, w^k_h} \mathbb{E}_{(x, y) \sim \mathcal{D}^k}\left[\ell_{k}(w_g, w^k_s, w^k_h; \;  x, y,\hat{y}_h)\right],
\end{equation}
locally for every client $k=1,\dots,n$. Here, $\ell_k$ is the loss from predicting $\hat{y}$ for the label $y$ given the input $x$ with the model from \eqref{mixture} over the data distribution $\mathcal{D}^k$ of client $k$. We freeze the weights of $w_g$, and only update $w_s^k$ and $w_h^k$.
A summary of the method is described in Algorithm \ref{algorithm}.

\begin{algorithm}[h]
  \caption{}
  \label{algorithm}
  \begin{algorithmic}[1] 
    \STATE{\bfseries input:} Models participating in \textsc{FedAvg} $w_1,\dots,w_k$, local gate $w_h^k$, learning rate $\eta$, decay rates $\beta_1, \beta_2$
    \STATE Initialize $w_1,\dots,w_k$ with the same random seed.
    \STATE Initialize $w_h^k$. 
    \STATE $w_g \gets$ \textsc{FedAvg}$(w_1,\dots,w_k) \;$ \textit{// Train for $E$ local epochs and $G$ communication rounds}
    \FOR{client $k$}
        \STATE Initialize specialist model $w_s^k \gets w_g$.
        \STATE $w_s^k \gets \text{Adam}(w_s^k, lr, \beta_1, \beta_2) \;$ \textit{// Fine-tune $w_g$ on each client $k$}
        \STATE Freeze global parameters $w_g$.
        \STATE $w_s^k, w_h^k \gets \text{Adam}(w_g, w_s^k, w_h^k , lr, \beta_1, \beta_2) \;$ \textit{// Train mixture of experts on client $k$}
    \ENDFOR
    \STATE{\bfseries output:} Trained mixture of experts: global model $w_g$, local experts $w_s^k$ and local gating functions $w_h^k$.
  \end{algorithmic}
\end{algorithm}
\section{Experimental setup}
We use two different ways of sampling client data to simulate heterogeneous distributions. The first setup is a more generalized version of the pathological non-iid setup as described in \citep{mcmahan2017communication} where each client is only assigned 2 classes. The second sampling strategy is performed using the Dirichlet distribution as described in \cite{yurochkin2019bayesian,hsu2019measuring}.

\textbf{Datasets and models.} Our experiments are carried out using two model architectures on three datasets. The dataset used are CIFAR-10 \citep{krizhevsky2009learning}, Fashion-MNIST \citep{xiao2017/online}, and AG News \citep{gulli2004ag}.

The CIFAR-10 dataset consists of 60 000 32x32 color images in 10 classes, with 6000 images per class. The dataset is split into 50 000 training images and 10 000 test images. 

The Fashion-MNIST dataset contains 70 000 28x28 gray-scale images of Zalando clothing in 10 classes. It is split into 60 000 training images and 10 000 test images.

The AG News topic classification dataset consists of 4 classes, each of which contains 30 000 training samples and 1 900 testing samples. In total there are 120 000 training samples and 7 600 testing samples.

For CIFAR-10 and Fashion-MNIST, the specialist model $f_s$ and the global model $f_g$ are CNNs with the same architecture. The CNN has two convolutional layers, each with a kernel size of 5 (the first with 6 channels, the second with 16), and two fully-connected layers with 120 and 84 units, respectively, with ReLU activations. This is followed by an output layer with a softmax activation. The gating function $h_k$ has the same architecture as $f_g$ and $f_s$, but with a sigmoid activation in the output layer instead of a softmax.

For AG News, both the local and the global models consist of an embedding layer with a dimension size of 100, a bi-directional LSTM layer with 64 nodes followed by an output layer with a softmax activation. The gating function has the same architecture in this case as well, but with a sigmoid activation in the output layer. We use the Adam optimizer \citep{kingma2014adam} to train all models.

\textbf{Pathological non-iid sampling.} The first way we create a skewed non-iid dataset for each client is by constructing a subset for each client with oversampling of specific classes. Sampling is performed such that the dataset of each client contains two majority classes which together form a fraction $p$ of the client data and the remaining classes form a fraction $(1-p)$ of the client data. We perform experiments with $p=\{0.2,0.3,\dots,1.0\}$ to see what effect the degree of heterogeneity has on performance. In the extreme case $p=1.0$, each client dataset only contains two classes in total, which is the same pathological non-iid setup as used for the MNIST dataset in \citep{mcmahan2017communication}. A majority class fraction of $p=0.2$ represents an iid setting for CIFAR-10 and Fashion-MNIST. For the AG News dataset which only has four classes, a fraction of $p=0.5$ represents an iid setting.

\textbf{Dirichlet distribution non-iid.} The second sampling strategy is to use the Dirichlet distribution as described in \cite{yurochkin2019bayesian,hsu2019measuring}. For each class we sample $\mathbf{n}_k\sim \text{Dir}_j(\alpha)$ and assign each client $j$ a proportion of $\mathbf{n}_{k,j}$ for class $k$. When $\alpha\to\infty$ we have an iid setting of equal number of instances per class for each client. When $\alpha\to 0$, we have a completely non-iid setting where each client dataset only has one class in total. We form experiments with $\alpha = \{0.05, 0.1, 0.5, 1.0, 10,100\}$.

\textbf{Opt-out factor and privacy.} Some users might want to opt out from participating to a global model, due to privacy reasons. These users will still receive the global model. To simulate this scenario in the experimental evaluation, we introduce an \textit{opt-out factor} denoted by $q$. This is a fraction deciding the number of clients participating in the \textsc{FedAvg} optimization. The clients that participate in the federated learning optimization have all their data in $\mathcal{D}_{\mathcal{I}}^k$, while the clients that opt out have all their data in $\mathcal{D}_{\mathcal{O}}^k$. $q=0$ means all clients are participating. We perform experiments varying $q$, to see how robust our algorithm is to different levels of client participation. In Figure \ref{fig:overview} we visualize how the opt-out factor can be used.

\textbf{\textsc{FedAvg} parameters.} For CIFAR-10 and Fashion-MNIST, the training is performed using 100 clients with 100 training samples per client. A client sampling fraction of $0.05$ is used, meaning that $5$ clients participate in each communication round. If the opt-out fraction $q$ is larger than $0$, we change the sampling fraction such that there always are $5$ clients that participate in every communication round. 

For AG News, we set the number of clients to 1000, with 100 training samples per client. We use a client sampling fraction of $0.05$, meaning that $50$ clients participate in each communication round. If the opt-out fraction $q$ is larger than $0$, we change the sampling fraction such that there always are $50$ clients that participate in every communication round. 

For all datasets we set the number of communication rounds to $1250$, number of local epochs to 3 and local batch size to $10$. We use early stopping and validate the performance of \textsc{FedAvg} on each participating client's local validation set every 50th communication round. The global model with the best mean validation loss over participating clients is returned.

\textbf{Baselines.} We use three different models as baselines. First, a locally trained model for every client, only trained on each clients own dataset. Second, \textsc{FedAvg}. Third, the final model output from \textsc{FedAvg} fine-tuned for each client on its own local data, denoted by $f^k_s$. We train the local model, the fine-tuned model and the mixture using early stopping for 500 epochs, monitoring local validation loss on each client and return the best performing model in each case.

\textbf{Evaluation.} We evaluate using both a \textit{local} (skewed) and a \textit{global} (balanced) test set. Each client has a local test set ($n=500$ samples) that mirrors its local data distribution. This test set is used to measure how well a model specializes to a client. The global test set ($n=1000$ samples) is a balanced test set (it contains the same number of data points for all classes) and is the same for all clients. We use this to measure how well a model generalizes. During evaluation, we sample 20 random clients and calculate the local and global test accuracies for all baselines and report a mean over the clients. All experiments were performed on a Tesla V100-SXM2-32GB, and all reported results are means over four runs.
\begin{table*}[t]
\centering
\caption{Accuracy on a global and local test set for all baselines on all datasets with varying majority class fractions $p$. Best performing specialist in bold. Opt-out fraction $q=0$. All results reported are over four runs.}
\label{tab_results}
\subfloat[CIFAR-10: Global test set]{\begin{tabular}{ccccc}
\toprule
$p$ &  FedAvg &  Local &  Fine-tuned &  Mixture  \\
\midrule
0.3 &            42.65 &           19.17 &        39.78 &         \textbf{41.57}\\
0.6 &            30.47 &           14.74 &        25.87 &         \textbf{26.37} \\
0.7 &            22.90 &           14.62 &        20.66 &         \textbf{21.45} \\
0.8 &            20.00 &           13.98 &        18.85 &         \textbf{19.55} \\
1.0 &            15.08 &           14.18 &        14.55 &         \textbf{14.66} \\
\bottomrule
\end{tabular}\label{global_cifar}}
\qquad
\subfloat[CIFAR-10: Local test set]{\begin{tabular}{ccccc}
\toprule
$p$ &  FedAvg &  Local &  Fine-tuned &  Mixture \\
\midrule
0.3 &                 43.32 &                23.39 &             42.98 &              \textbf{43.86} \\
0.6 &                 30.13 &                42.19 &             \textbf{50.97} &              50.57 \\
0.7 &                 21.81 &                50.23 &             \textbf{54.98} &              54.73 \\
0.8 &                 16.78 &                59.89 &             \textbf{64.54} &              64.47 \\
1.0 &                 13.62 &                77.00 &             78.32 &              \textbf{78.56}\\
\bottomrule
\end{tabular}\label{local_cifar}}
\qquad
\subfloat[Fashion-MNIST: Global test set]{\begin{tabular}{ccccc}
\toprule
$p$ &  FedAvg &  Local &  Fine-tuned &  Mixture \\
\midrule
0.3 &            71.12 &           40.65 &        69.32 &         \textbf{70.50} \\
0.6 &            66.85 &           18.07 &        61.43 &         \textbf{64.82} \\
0.7 &            64.45 &           17.57 &        58.95 &         \textbf{62.15} \\
0.8 &            67.45 &           17.69 &        58.66 &         \textbf{61.53} \\
1.0 &            49.43 &           18.11 &        \textbf{23.69} &         22.64 \\
\bottomrule
\end{tabular}\label{global_fashion}}
\qquad
\subfloat[Fashion-MNIST: Local test set]{
\begin{tabular}{lrrrr}
\toprule
$p$ &  FedAvg &  Local &  Fine-tuned &  Mixture \\
\midrule
0.3 &                 70.66 &                45.77 &             69.76 &              \textbf{70.42} \\
0.6 &                 65.14 &                55.16 &             71.01 &              \textbf{71.18} \\
0.7 &                 64.67 &                65.12 &             \textbf{74.86} &              74.63 \\
0.8 &                 66.45 &                74.84 &             76.02 &              \textbf{76.70} \\
1.0 &                 48.01 &                \textbf{94.22} &             91.28 &              92.10 \\
\bottomrule
\end{tabular}\label{local_fashion}}
\qquad
\subfloat[AG News: Global test set]{
\begin{tabular}{ccccc}
\toprule
$p$ &  FedAvg &  Local &  Fine-tuned &  Mixture \\
\midrule
0.5 &            80.31 &           28.82 &        75.51 &         \textbf{78.03} \\
0.7 &            10.42 &            3.88 &         9.79 &         \textbf{10.15 }\\
0.8 &            10.39 &            3.82 &         9.59 &         \textbf{10.01} \\
0.9 &             9.73 &            3.91 &         8.90 &          \textbf{9.43} \\
1.0 &             8.62 &            3.95 &         6.43 &          \textbf{7.29} \\
\bottomrule
\end{tabular}\label{global_ag}}
\qquad
\subfloat[AG News: Local test set]{\begin{tabular}{ccccc}
\toprule
$p$ &  FedAvg &  Local &  Fine-tuned &  Mixture \\
\midrule
0.5 &                 81.93 &                34.22 &             78.48 &              \textbf{80.53} \\
0.7 &                 79.23 &                48.92 &             80.37 &              \textbf{81.98} \\
0.8 &                 75.45 &                56.21 &             81.49 &              \textbf{82.66} \\
0.9 &                 68.47 &                63.71 &             \textbf{83.44} &              82.96 \\
1.0 &                 54.86 &                72.34 &             \textbf{87.26} &              86.51 \\
\bottomrule
\end{tabular}\label{local_ag}}
\end{table*}
\section{Results and discussion}
For the sake of reproducibility, all code is made available. \footnote{\texttt{Link to github repo will be made available here.}}

\begin{figure*}[t]
    \centering
    \subfloat[CIFAR-10]{\includegraphics[width=0.33\textwidth]{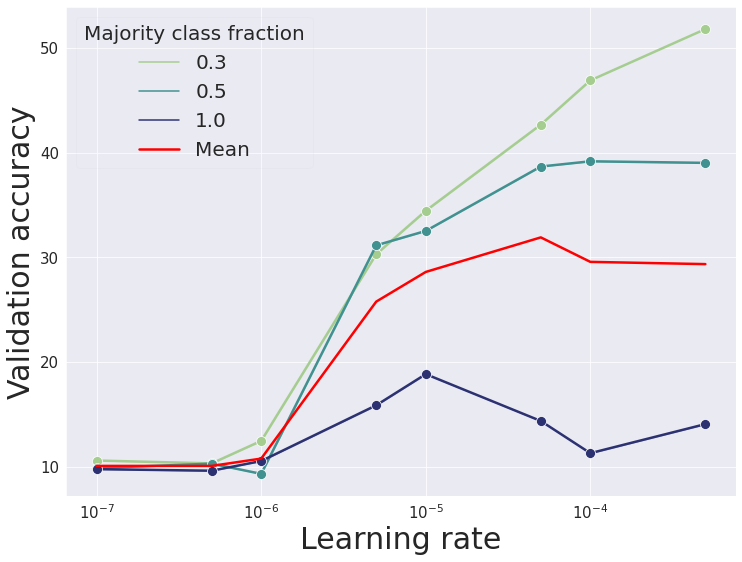}\label{cifar_lr}} 
    \subfloat[Fashion-MNIST]{\includegraphics[width=0.33\textwidth]{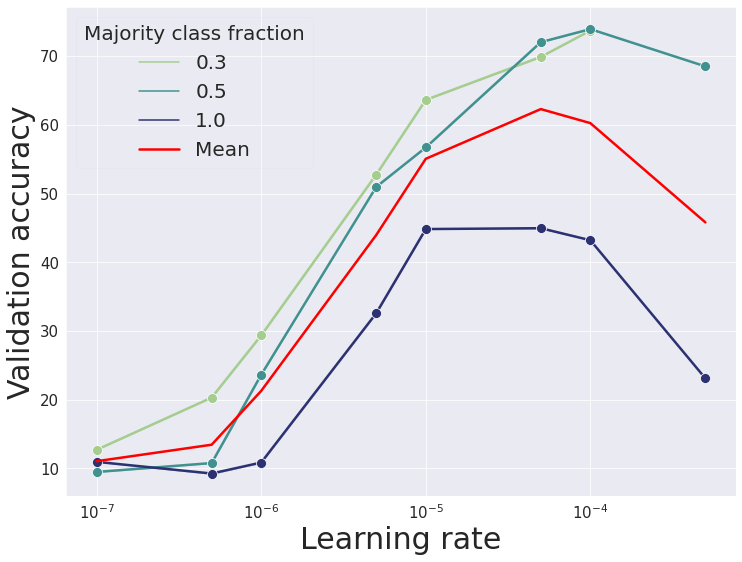}\label{fashion_lr}}
    \subfloat[AG News]{\includegraphics[width=0.33\textwidth]{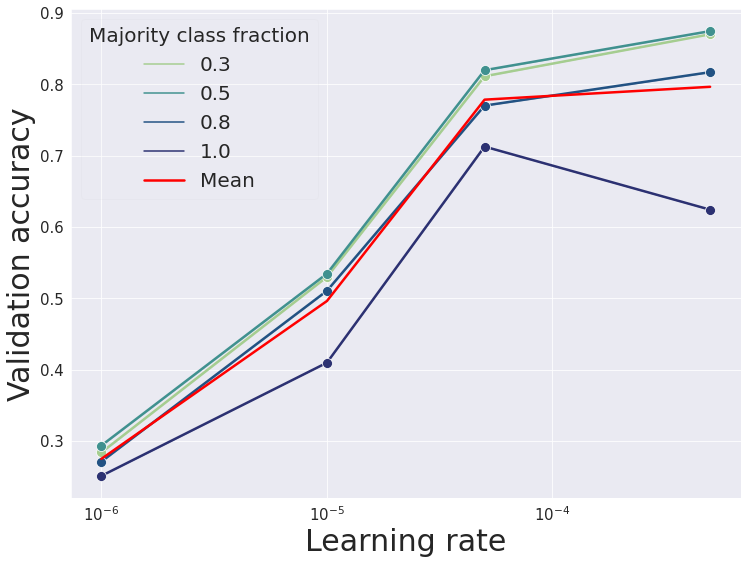}\label{ag_lr}}
    \caption{Learning rate vs balanced validation accuracy for \textsc{FedAvg} on (a) CIFAR-10, (b) Fashion-MNIST and (c) AG News using different majority class fractions $p$. Reported values are means over four runs.}
    \label{fig:lr}
\end{figure*}
\begin{figure*}[t]
    \centering
    \subfloat[CIFAR-10 (100 clients)]{\includegraphics[width=0.33\textwidth]{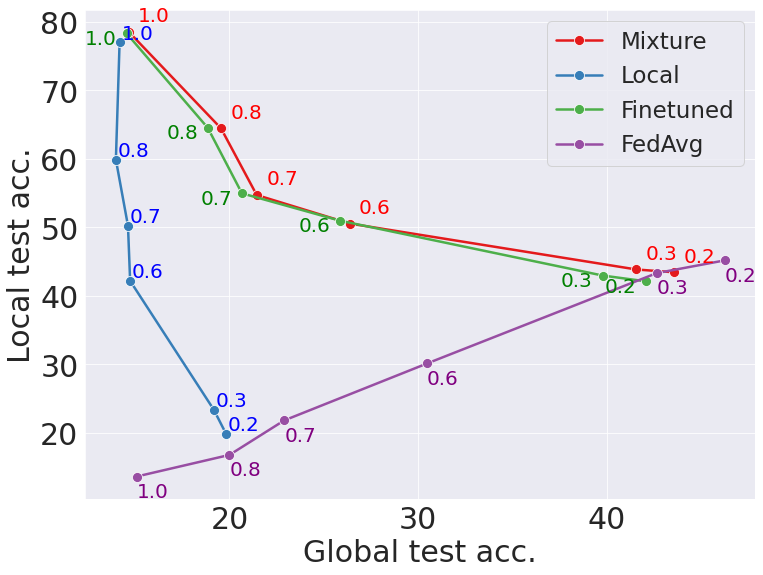}}
    \subfloat[Fashion-MNIST (100 clients)]{\includegraphics[width=0.33\textwidth]{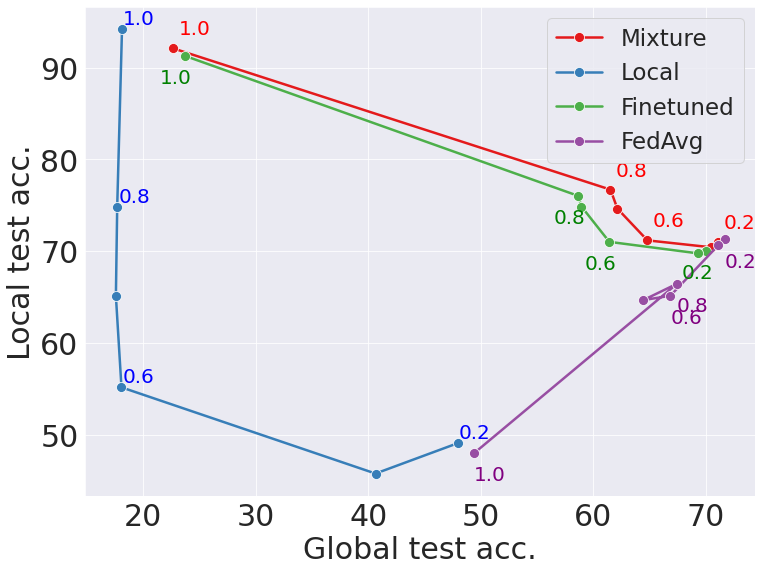}}
    \subfloat[AG News (1000 clients)]{\includegraphics[width=0.33\textwidth]{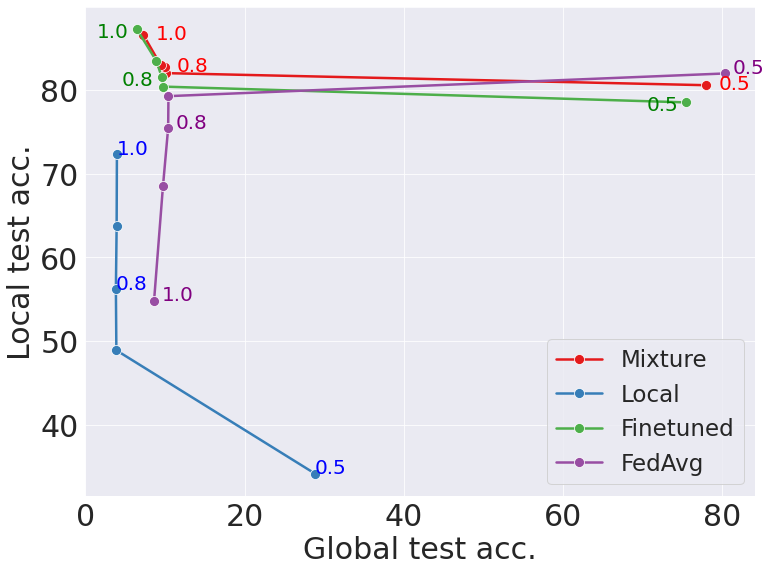}}
    \caption{Accuracy on balanced test data ($x$-axis) vs unbalanced local test data ($y$-axis) for the three datasets with opt-out fraction $q=0$. Different majority class fractions $p$ are shown as colored numbers. Reported values are means over four runs.}
    \label{fig:cifaropt}
\end{figure*}
Figure \ref{fig:lr} shows a learning rate sweep for \textsc{FedAvg} on all three datasets using different majority class fractions $p$. The sweep was carried out over learning rates $\eta=\{10^{-7},5\cdot10^{-7},\dots,10^{-3},5\cdot10^{-3}\}$. The accuracy was calculated on a balanced validation set. The learning rate of $\eta = 5\cdot10^{-5}$ yielded the best validation accuracy for both CIFAR-10 and Fashion-MNIST, and given these results, we use this learning rate for training \textsc{FedAvg} in all experiments for these two datasets. For the AG News dataset best overall performing learning rate was found to be $\eta=5\cdot10^{-4}$.

The same learning rates of $\eta = 5\cdot10^{-5}$ (CIFAR-10 and Fashion-MNIST) and $\eta=5\cdot10^{-4}$ (AG News) was set to train the local baseline models. For the fine-tuned baseline model and the mixture of experts, a lower learning rate was used of $\eta = 10^{-5}$ for CIFAR-10 and Fashion-MNIST and $\eta = 10^{-6}$ for AG News. In the appendix we show training and validation losses for \textsc{FedAvg} over communication rounds.

In Table \ref{tab_results} we see results for the three datasets for varying majority class fractions $p$. The leftmost Tables \ref{global_cifar}, \ref{global_fashion} and \ref{global_ag} show the results for a global (balanced) dataset, which is the same for all clients. The rightmost tables \ref{local_cifar}, \ref{local_fashion} and \ref{local_ag} show the results on a local (unbalanced) dataset mirroring each clients distribution. In bold we present the best performing specialist. We note here that our proposed mixture of experts is overall the best specialist model in terms of generalization, whereas it performs roughly equally as good as the fine-tuned specialist on a local test set. 

This is further visualized for all datasets in Figure \ref{fig:cifaropt}. Here the global test accuracy is shown on the $x$-axis and the local test accuracy is shown on the $y$-axis for the different baselines. We note that the mixture of expert consistently outperforms the fine-tuned specialist on all three datasets, performing roughly equal on a local test set while consistently outperforming it on the global test set.

\begin{figure*}[t]
    \centering
    \subfloat[CIFAR-10 (100 clients)]{\includegraphics[width=0.33\textwidth]{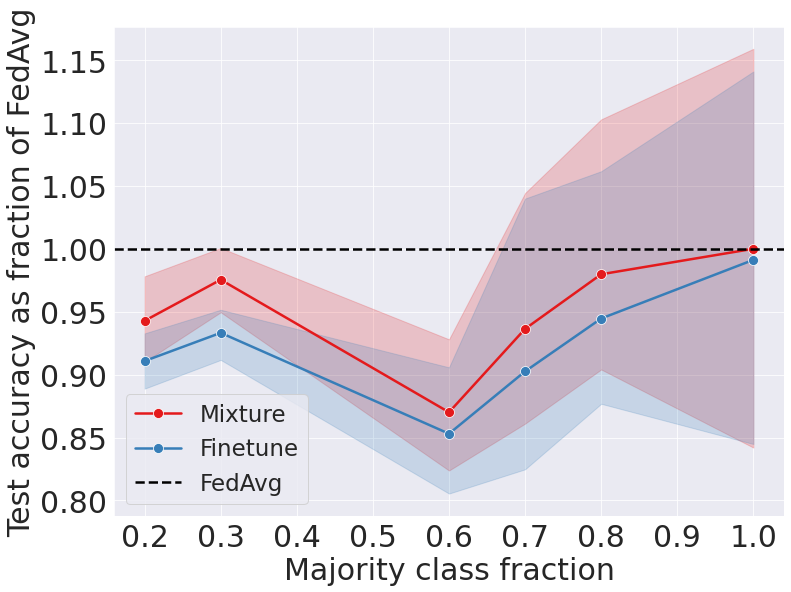}\label{cifar_frac}}
    \subfloat[AG News (1000 clients)]{\includegraphics[width=0.33\textwidth]{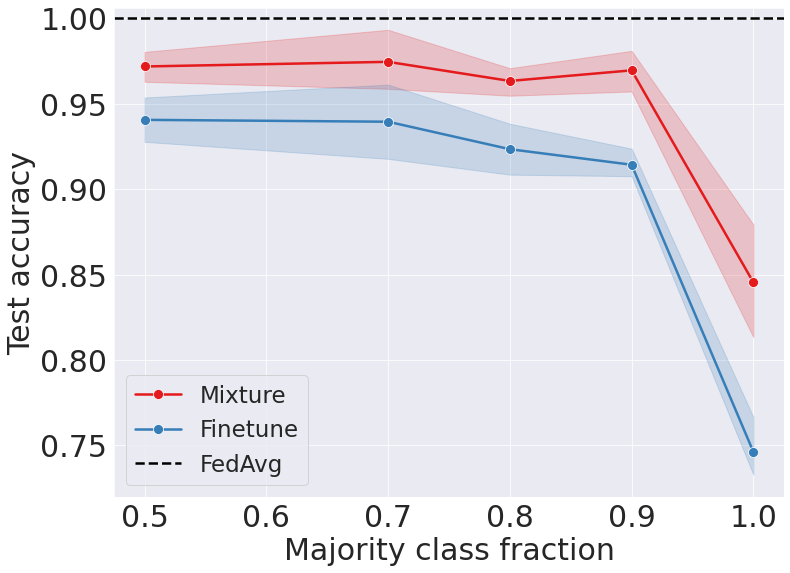}\label{ag_frac}} 
    \subfloat[Fashion-MNIST (100 clients)]{\includegraphics[width=0.33\textwidth]{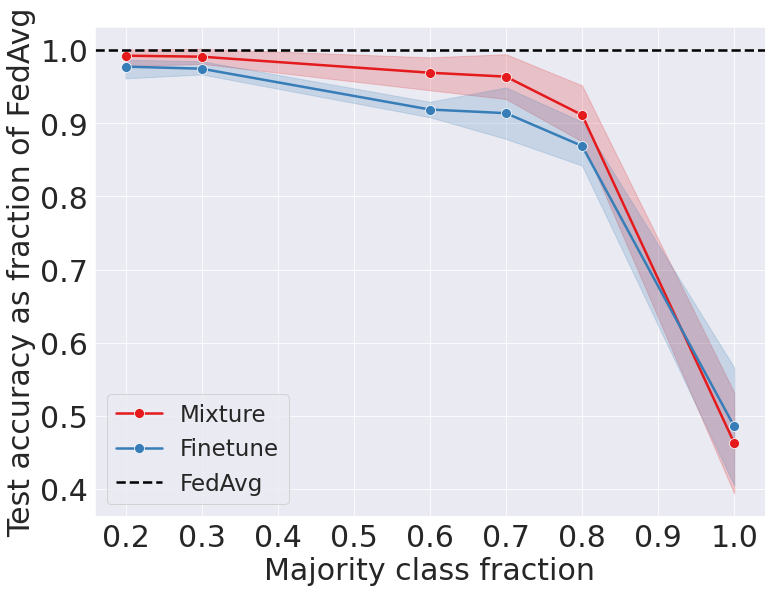}\label{fashion_frac}}
    \caption{Test accuracy with a $95\%$ confidence interval vs majority class fractions for fine-tuned baseline and the mixture on a global (balanced) test set for (a) CIFAR-10 and (b) AG News and (c) Fashion-MNIST, as a fraction of \textsc{FedAvg} test accuracy. Opt-out fraction $q=0$.  Reported values are means over four runs.}
    \label{fig:fractions}
\end{figure*}

\begin{figure*}[t]
    \centering
    \subfloat[CIFAR-10 (100 clients)]{\includegraphics[width=0.33\textwidth]{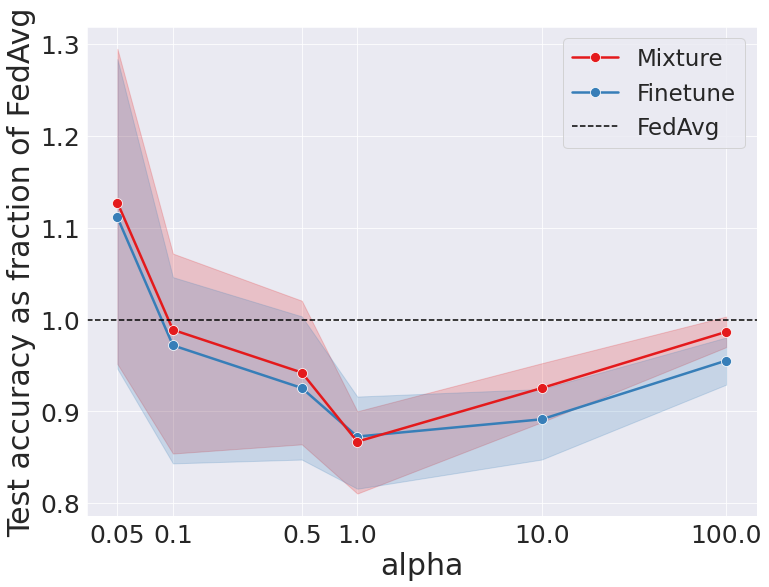}\label{cifar_alpha}}
    \subfloat[Fashion-MNIST (100 clients)]{\includegraphics[width=0.33\textwidth]{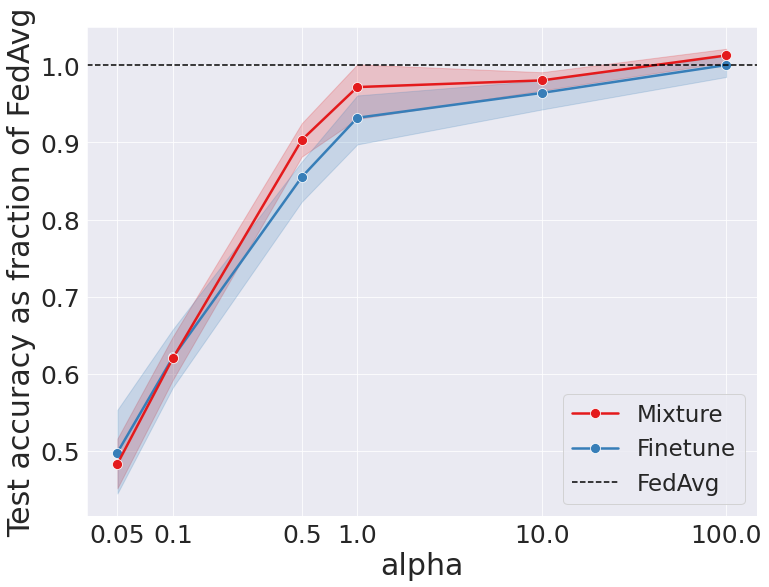}\label{fashion_alpha}} 
    \caption{Test accuracy with a $95\%$ confidence interval vs Dirichlet parameter $\alpha$ for fine-tuned baseline and the mixture on a global (balanced) test set for (a) CIFAR-10 and (b) Fashion-MNIST, as a fraction of \textsc{FedAvg} test accuracy. Opt-out fraction $q=0$.  Reported values are means over four runs.}
    \label{fig:alpha_fractions}
\end{figure*}

In Figure \ref{fig:fractions} global test accuracies for the fine-tuned baseline and the mixture of experts on all datasets are shown, as a fraction of \textsc{FedAvg} test accuracy. Here we see that that our proposed method consistently outperforms the fine-tuned baseline in terms of generalization, being the closest to \textsc{FedAvg} performance in all settings. In Figure \ref{fig:alpha_fractions} similar results for different Dirichlet $\alpha$ values are presented, instead of majority class fractions $p$. 

\begin{figure}[H]
    \centering
    \subfloat[Opt out-fraction $q=0.9$]{\includegraphics[width=0.33\textwidth]{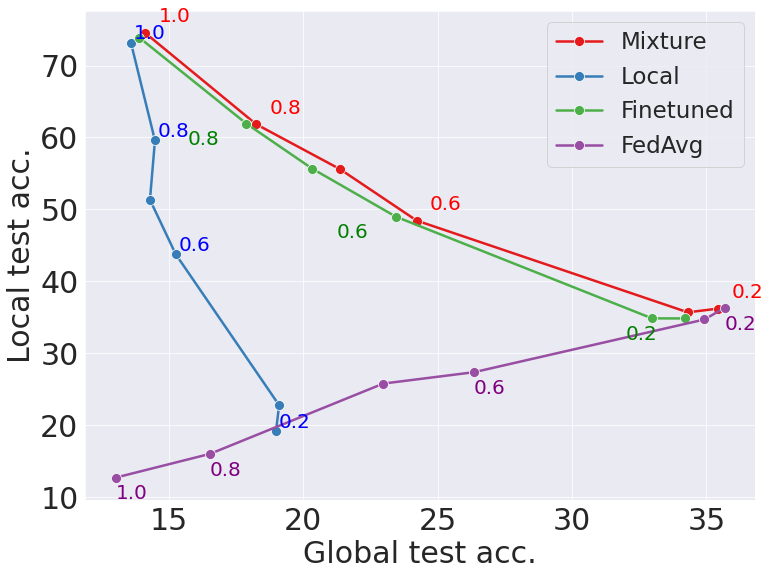}\label{cifar_9}} \\
    \subfloat[Opt-out fraction $q=0.95$]{\includegraphics[width=0.33\textwidth]{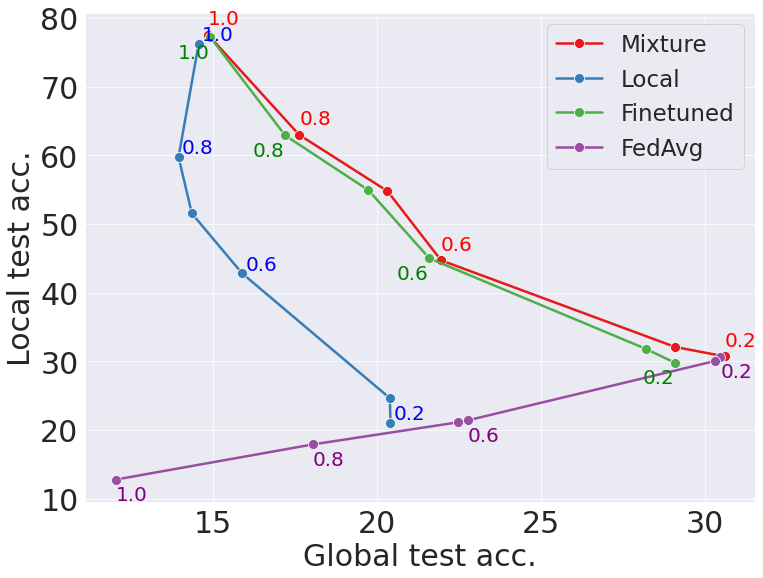}\label{cifar_95}}
    \caption{Test accuracy on a global test set ($x$-axis) and local test set ($y$-axis) for the CIFAR-10 dataset, with two different opt-out fractions $q$. Majority class fractions are shown in colored numbers.}
    \label{fig:opt_out}
\end{figure}

\newpage
\textbf{Opt-out fractions.} Experiments were carried out to test what effect client opt-out has on performance. The results can be seen in Figure \ref{fig:opt_out} for CIFAR-10 over varying majority class fractions $p$ with large opt-out fractions of $q=\{0.9,0.95\}$, meaning that $90\%$ and $95\%$ of clients, respectively, choose not to participate in the federated learning, but still obtains the global model at the end of training. Similar to the results with no opt-out ($q=0$), our proposed model outperforms the fine-tuned baseline on the global test set, while performing on par on the local test sets. In Figure \ref{cifar_95}, we see that in the iid case of $p=0.2$ that the mixture not only outperforms the fine-tuned model, but also \textsc{FedAvg} in both generalization and specialization. This shows that the mixture in this setting is more robust to many clients opting out from federated learning.
\section{Conclusions}
To address the problems of learning a personalized model in a federated setting when the client data is heterogeneous, we have proposed a novel framework for federated mixtures of experts where a global model is combined with a local specialist model. 
We find that by combining the two expert models we achieve high performance on local client datasets, with minimal loss on generalization as compared to a fine-tuned baseline on two image classification datasets and one text classification dataset in highly non-iid settings, and achieve test accuracies on par with \textsc{FedAvg} in iid settings.

Our approach is not only an intuitive approach for the generalist vs specialist balance, but also allows for varying level of participation of the different clients in the federation. As such, the framework gives strong privacy guarantees, where clients who do not want to disclose their data are able to opt out and keep their data completely private. The experiments show that our proposed solution is robust to a high opt-out fraction of users, as seen in Figure \ref{fig:cifaropt}. It thus constitutes a flexible solution for strong privacy guarantees in real-world settings where users might not want to disclose their model to a central server.

The proposed framework is compatible with any gradient-based machine learning model, and can incorporate combinations of these, strengthening the potential of this direction of research, and leveraging the beneficial properties of ensembles of various machine learning models.

In this work we limited our experiments to non-identical distributions in the form of prior probability shift. We hypothesize that our proposed method can handle other distributional shifts, such as covariate or concept shifts as well, and see this as an interesting direction for future work.


\bibliography{bibliography.bib}

\begin{thebibliography}{31}
\providecommand{\natexlab}[1]{#1}
\providecommand{\url}[1]{\texttt{#1}}
\expandafter\ifx\csname urlstyle\endcsname\relax
  \providecommand{\doi}[1]{doi: #1}\else
  \providecommand{\doi}{doi: \begingroup \urlstyle{rm}\Url}\fi

\bibitem[Arivazhagan et~al.(2019)Arivazhagan, Aggarwal, Singh, and
  Choudhary]{arivazhagan2019federated}
Arivazhagan, M.~G., Aggarwal, V., Singh, A.~K., and Choudhary, S.
\newblock Federated learning with personalization layers.
\newblock \emph{arXiv preprint arXiv:1912.00818}, 2019.

\bibitem[Bellet et~al.(2018)Bellet, Guerraoui, Taziki, and
  Tommasi]{bellet2018personalized}
Bellet, A., Guerraoui, R., Taziki, M., and Tommasi, M.
\newblock Personalized and private peer-to-peer machine learning.
\newblock In \emph{International Conference on Artificial Intelligence and
  Statistics}, pp.\  473--481, 2018.

\bibitem[Deng et~al.(2020)Deng, Kamani, and Mahdavi]{deng2020adaptive}
Deng, Y., Kamani, M.~M., and Mahdavi, M.
\newblock Adaptive personalized federated learning.
\newblock \emph{arXiv preprint arXiv:2003.13461}, 2020.

\bibitem[Fallah et~al.(2020)Fallah, Mokhtari, and
  Ozdaglar]{fallah2020personalized}
Fallah, A., Mokhtari, A., and Ozdaglar, A.
\newblock Personalized federated learning: A meta-learning approach.
\newblock \emph{arXiv preprint arXiv:2002.07948}, 2020.

\bibitem[Finn et~al.(2017)Finn, Abbeel, and Levine]{finn2017model}
Finn, C., Abbeel, P., and Levine, S.
\newblock Model-agnostic meta-learning for fast adaptation of deep networks.
\newblock In \emph{ICML}, 2017.

\bibitem[Gulli(2004)]{gulli2004ag}
Gulli, A.
\newblock Ag’s corpus of news articles.
\newblock \emph{URL http://groups. di. unipi. it/\~{} gulli/AG\_
  corpus\_of\_news\_articles. html}, 2004.

\bibitem[Hanzely \& Richt{\'a}rik(2020)Hanzely and
  Richt{\'a}rik]{hanzely2020federated}
Hanzely, F. and Richt{\'a}rik, P.
\newblock Federated learning of a mixture of global and local models.
\newblock \emph{arXiv preprint arXiv:2002.05516}, 2020.

\bibitem[Hard et~al.(2018)Hard, Kiddon, Ramage, Beaufays, Eichner, Rao,
  Mathews, and Augenstein]{hard2018federated}
Hard, A., Kiddon, C.~M., Ramage, D., Beaufays, F., Eichner, H., Rao, K.,
  Mathews, R., and Augenstein, S.
\newblock Federated learning for mobile keyboard prediction, 2018.
\newblock URL \url{https://arxiv.org/abs/1811.03604}.

\bibitem[He et~al.(2020)He, Avestimehr, and Annavaram]{he2020group}
He, C., Avestimehr, S., and Annavaram, M.
\newblock Group knowledge transfer: Collaborative training of large cnns on the
  edge.
\newblock \emph{Advances in Neural Information Processing Systems 33
  proceedings (NeurIPS)}, 2020.

\bibitem[Hsieh et~al.(2020)Hsieh, Phanishayee, Mutlu, and
  Gibbons]{hsieh2020non}
Hsieh, K., Phanishayee, A., Mutlu, O., and Gibbons, P.
\newblock The non-iid data quagmire of decentralized machine learning.
\newblock In \emph{International Conference on Machine Learning}, pp.\
  4387--4398. PMLR, 2020.

\bibitem[Hsu et~al.(2019)Hsu, Qi, and Brown]{hsu2019measuring}
Hsu, T.-M.~H., Qi, H., and Brown, M.
\newblock Measuring the effects of non-identical data distribution for
  federated visual classification.
\newblock \emph{arXiv preprint arXiv:1909.06335}, 2019.

\bibitem[Jacobs et~al.(1991)Jacobs, Jordan, Nowlan, and
  Hinton]{jacobs1991adaptive}
Jacobs, R.~A., Jordan, M.~I., Nowlan, S.~J., and Hinton, G.~E.
\newblock Adaptive mixtures of local experts.
\newblock \emph{Neural computation}, 3\penalty0 (1):\penalty0 79--87, 1991.

\bibitem[Jeong et~al.(2018)Jeong, Oh, Kim, Park, Bennis, and
  Kim]{jeong2018communication}
Jeong, E., Oh, S., Kim, H., Park, J., Bennis, M., and Kim, S.-L.
\newblock Communication-efficient on-device machine learning: Federated
  distillation and augmentation under non-iid private data.
\newblock \emph{arXiv preprint arXiv:1811.11479}, 2018.

\bibitem[Jiang et~al.(2019)Jiang, Kone{\v{c}}n{\`y}, Rush, and
  Kannan]{jiang2019improving}
Jiang, Y., Kone{\v{c}}n{\`y}, J., Rush, K., and Kannan, S.
\newblock Improving federated learning personalization via model agnostic meta
  learning.
\newblock \emph{arXiv preprint arXiv:1909.12488}, 2019.

\bibitem[Johnson \& Khoshgoftaar(2019)Johnson and
  Khoshgoftaar]{johnson2019survey}
Johnson, J.~M. and Khoshgoftaar, T.~M.
\newblock Survey on deep learning with class imbalance.
\newblock \emph{Journal of Big Data}, 6\penalty0 (1):\penalty0 27, 2019.

\bibitem[Kairouz et~al.(2019)Kairouz, McMahan, Avent, Bellet, Bennis, Bhagoji,
  Bonawitz, Charles, Cormode, Cummings, et~al.]{kairouz2019advances}
Kairouz, P., McMahan, H.~B., Avent, B., Bellet, A., Bennis, M., Bhagoji, A.~N.,
  Bonawitz, K., Charles, Z., Cormode, G., Cummings, R., et~al.
\newblock Advances and open problems in federated learning.
\newblock \emph{arXiv preprint arXiv:1912.04977}, 2019.

\bibitem[Kingma \& Ba(2014)Kingma and Ba]{kingma2014adam}
Kingma, D.~P. and Ba, J.
\newblock Adam: A method for stochastic optimization.
\newblock \emph{arXiv preprint arXiv:1412.6980}, 2014.

\bibitem[Kone{\v{c}}n{\`y} et~al.(2016)Kone{\v{c}}n{\`y}, McMahan, Yu,
  Richt{\'a}rik, Suresh, and Bacon]{konevcny2016federated}
Kone{\v{c}}n{\`y}, J., McMahan, H.~B., Yu, F.~X., Richt{\'a}rik, P., Suresh,
  A.~T., and Bacon, D.
\newblock Federated learning: Strategies for improving communication
  efficiency.
\newblock \emph{arXiv preprint arXiv:1610.05492}, 2016.

\bibitem[Krizhevsky et~al.(2009)Krizhevsky, Hinton,
  et~al.]{krizhevsky2009learning}
Krizhevsky, A., Hinton, G., et~al.
\newblock Learning multiple layers of features from tiny images.
\newblock 2009.

\bibitem[Lin et~al.(2020)Lin, Kong, Stich, and Jaggi]{lin2020ensemble}
Lin, T., Kong, L., Stich, S.~U., and Jaggi, M.
\newblock Ensemble distillation for robust model fusion in federated learning.
\newblock \emph{Advances in Neural Information Processing Systems}, 33, 2020.

\bibitem[Mansour et~al.(2009)Mansour, Mohri, and
  Rostamizadeh]{mansour2009domain}
Mansour, Y., Mohri, M., and Rostamizadeh, A.
\newblock Domain adaptation: Learning bounds and algorithms.
\newblock \emph{arXiv preprint arXiv:0902.3430}, 2009.

\bibitem[McMahan et~al.(2017)McMahan, Moore, Ramage, Hampson, and
  y~Arcas]{mcmahan2017communication}
McMahan, B., Moore, E., Ramage, D., Hampson, S., and y~Arcas, B.~A.
\newblock Communication-efficient learning of deep networks from decentralized
  data.
\newblock In \emph{Artificial Intelligence and Statistics}, pp.\  1273--1282.
  PMLR, 2017.

\bibitem[Oquab et~al.(2014)Oquab, Bottou, Laptev, and Sivic]{oquab2014learning}
Oquab, M., Bottou, L., Laptev, I., and Sivic, J.
\newblock Learning and transferring mid-level image representations using
  convolutional neural networks.
\newblock In \emph{Proceedings of the IEEE conference on computer vision and
  pattern recognition}, pp.\  1717--1724, 2014.

\bibitem[Peterson et~al.(2019)Peterson, Kanani, and
  Marathe]{peterson2019private}
Peterson, D., Kanani, P., and Marathe, V.~J.
\newblock Private federated learning with domain adaptation.
\newblock \emph{arXiv preprint arXiv:1912.06733}, 2019.

\bibitem[Shokri \& Shmatikov(2015)Shokri and Shmatikov]{shokri2015privacy}
Shokri, R. and Shmatikov, V.
\newblock Privacy-preserving deep learning.
\newblock In \emph{Proceedings of the 22nd ACM SIGSAC conference on computer
  and communications security}, pp.\  1310--1321, 2015.

\bibitem[Vanhaesebrouck et~al.(2017)Vanhaesebrouck, Bellet, and
  Tommasi]{vanhaesebrouck2016decentralized}
Vanhaesebrouck, P., Bellet, A., and Tommasi, M.
\newblock Decentralized collaborative learning of personalized models over
  networks.
\newblock In \emph{Artificial Intelligence and Statistics}, pp.\  509--517.
  PMLR, 2017.

\bibitem[Wang et~al.(2019)Wang, Mathews, Kiddon, Eichner, Beaufays, and
  Ramage]{wang2019federated}
Wang, K., Mathews, R., Kiddon, C., Eichner, H., Beaufays, F., and Ramage, D.
\newblock Federated evaluation of on-device personalization.
\newblock \emph{arXiv preprint arXiv:1910.10252}, 2019.

\bibitem[{Wang} et~al.(2019){Wang}, {Song}, {Zhang}, {Song}, {Wang}, and
  {Qi}]{wang2019beyond}
{Wang}, Z., {Song}, M., {Zhang}, Z., {Song}, Y., {Wang}, Q., and {Qi}, H.
\newblock Beyond inferring class representatives: User-level privacy leakage
  from federated learning.
\newblock In \emph{IEEE INFOCOM 2019 - IEEE Conference on Computer
  Communications}, pp.\  2512--2520, 2019.

\bibitem[Xiao et~al.(2017)Xiao, Rasul, and Vollgraf]{xiao2017/online}
Xiao, H., Rasul, K., and Vollgraf, R.
\newblock Fashion-mnist: a novel image dataset for benchmarking machine
  learning algorithms, 2017.

\bibitem[Yurochkin et~al.(2019)Yurochkin, Agarwal, Ghosh, Greenewald, Hoang,
  and Khazaeni]{yurochkin2019bayesian}
Yurochkin, M., Agarwal, M., Ghosh, S., Greenewald, K., Hoang, N., and Khazaeni,
  Y.
\newblock Bayesian nonparametric federated learning of neural networks.
\newblock In \emph{International Conference on Machine Learning}, pp.\
  7252--7261. PMLR, 2019.

\bibitem[Zhao et~al.(2018)Zhao, Li, Lai, Suda, Civin, and
  Chandra]{zhao2018federated}
Zhao, Y., Li, M., Lai, L., Suda, N., Civin, D., and Chandra, V.
\newblock Federated learning with non-iid data.
\newblock \emph{arXiv preprint arXiv:1806.00582}, 2018.

\end{thebibliography}
\bibliographystyle{icml2021}


%



\end{document}


\twocolumn[
\icmltitle{Specialized federated learning using a mixture of experts}



\icmlsetsymbol{equal}{*}

\begin{icmlauthorlist}
\icmlauthor{Aeiau Zzzz}{equal,to}
\icmlauthor{Bauiu C.~Yyyy}{equal,to,goo}
\icmlauthor{Cieua Vvvvv}{goo}
\icmlauthor{Iaesut Saoeu}{ed}
\icmlauthor{Fiuea Rrrr}{to}
\icmlauthor{Tateu H.~Yasehe}{ed,to,goo}
\icmlauthor{Aaoeu Iasoh}{goo}
\icmlauthor{Buiui Eueu}{ed}
\icmlauthor{Aeuia Zzzz}{ed}
\icmlauthor{Bieea C.~Yyyy}{to,goo}
\icmlauthor{Teoau Xxxx}{ed}
\icmlauthor{Eee Pppp}{ed}
\end{icmlauthorlist}

\icmlaffiliation{to}{Department of Computation, University of Torontoland, Torontoland, Canada}
\icmlaffiliation{goo}{Googol ShallowMind, New London, Michigan, USA}
\icmlaffiliation{ed}{School of Computation, University of Edenborrow, Edenborrow, United Kingdom}

\icmlcorrespondingauthor{Cieua Vvvvv}{c.vvvvv@googol.com}
\icmlcorrespondingauthor{Eee Pppp}{ep@eden.co.uk}

\icmlkeywords{Machine Learning, ICML}

\vskip 0.3in
]



\printAffiliationsAndNotice{\icmlEqualContribution} 

\appendix
\section{\textsc{FedAvg} training details for CIFAR-10}

In Figures \ref{fig:train_loss}, \ref{fig:val_loss} and \ref{fig:val_acc} results for \textsc{FedAvg} training is shown for different majority class fractions $p$ on CIFAR-10: $p=1.0$ (green), $p=0.5$ (red) and $p=0.3$ (magenta). Training is stopped when validation loss is not improved upon after 160 communication rounds and the weights with the lowest validation loss are returned.

\begin{figure}[H]
    \centering
    \includesvg[scale=1.0]{fedAvg_train_loss.svg}
    \caption{Communication rounds vs training loss for \textsc{FedAvg} for three different majority class fractions: $p=1.0$ (green), $p=0.5$ (red) and $p=0.3$ (magenta)}
    \label{fig:train_loss}
\end{figure}

\begin{figure}[H]
    \centering
    \includesvg[scale=0.9]{fedAvg_val_loss.svg}
    \caption{Communication rounds vs validation loss for \textsc{FedAvg} for three different majority class fractions: $p=1.0$ (green), $p=0.5$ (red) and $p=0.3$ (magenta)}
    \label{fig:val_loss}
\end{figure}

\begin{figure}[H]
    \centering
    \includesvg[scale=0.9]{fedAvg_val_acc.svg}
    \caption{Communication rounds vs validation accuracy for \textsc{FedAvg} for three different majority class fractions: $p=1.0$ (green), $p=0.5$ (red) and $p=0.3$ (magenta)}
    \label{fig:val_acc}
\end{figure}
\section{\textsc{FedAvg} training details for Fashion-MNIST}
In Figures \ref{fig:train_loss_f}, \ref{fig:val_loss_f} and \ref{fig:val_acc_f} results for \textsc{FedAvg} training is shown for different majority class fractions $p$ on Fashion-MNIST: $p=1.0$ (blue), $p=0.5$ (magenta) and $p=0.3$ (green). Training is stopped when validation loss is not improved upon after 160 communication rounds and the weights with the lowest validation loss are returned.
\begin{figure}[H]
    \centering
    \includesvg[scale=1.0]{fedAvg_train_loss_fashion.svg}
    \caption{Communication rounds vs training loss for \textsc{FedAvg} for three different majority class fractions: $p=1.0$ (blue), $p=0.5$ (magenta) and $p=0.3$ (green)}
    \label{fig:train_loss_f}
\end{figure}

\begin{figure}[H]
    \centering
    \includesvg[scale=0.9]{fedAvg_val_loss_fashion.svg}
    \caption{Communication rounds vs validation loss for \textsc{FedAvg} for three different majority class fractions: $p=1.0$ (blue), $p=0.5$ (magenta) and $p=0.3$ (green)}
    \label{fig:val_loss_f}
\end{figure}

\begin{figure}[H]
    \centering
    \includesvg[scale=0.9]{fedAvg_val_acc_fashion.svg}
    \caption{Communication rounds vs validation accuracy for \textsc{FedAvg} for three different majority class fractions:$p=1.0$ (blue), $p=0.5$ (magenta) and $p=0.3$ (green)}
    \label{fig:val_acc_f}
\end{figure}
%

